\documentclass[runningheads]{llncs}
\usepackage[T1]{fontenc}
\usepackage{graphicx}
\usepackage{algorithmicx}
\usepackage{algorithm}
\usepackage[noend]{algpseudocode}
\usepackage{pdfpages}
\usepackage{caption}
\usepackage{subcaption}
\usepackage{multirow}

\def\NoNumber#1{{\def\alglinenumber##1{}\State #1}\addtocounter{ALG@line}{-1}}
\begin{document}
\title{Anomalous Client Detection in Federated Learning}
%
%
\author{Dipanwita Thakur\inst{1}\orcidID{0000-0003-2895-1425} \and
Antonella Guzzo\inst{1}\orcidID{0000-0003-3159-0536} \and
Giancarlo Fortino\inst{1}\orcidID{0000-0002-4039-891X}}
\authorrunning{D. Thakur et al.}
%
\institute{University of Calabria, Rende CS 87036, Italy \\
\email{\{dipanwita.thakur,antonella.guzzo,giancarlo.fortino\}@unical.it}}
\maketitle              
\begin{abstract}
Federated learning (FL), with the growing IoT and edge computing, is seen as a promising solution for applications that are latency- and privacy-aware. However, due to the widespread dispersion of data across many clients, it is challenging to monitor client anomalies caused by malfunctioning devices or unexpected events. The majority of FL solutions now in use concentrate on the classification problem, ignoring situations in which anomaly detection may also necessitate privacy preservation and effectiveness. The system in federated learning is unable to manage the potentially flawed behavior of its clients completely. These behaviors include sharing arbitrary parameter values and causing a delay in convergence since clients are chosen at random without knowing the malfunctioning behavior of the client. Client selection is crucial in terms of the efficiency of the federated learning framework. The challenges such as client drift and handling slow clients with low computational capability are well-studied in FL. However, the detection of anomalous clients either for security or for overall performance in the FL frameworks is hardly studied in the literature. In this paper, we propose an anomaly client detection algorithm to overcome malicious client attacks and client drift in FL frameworks. Instead of random client selection, our proposed method utilizes anomaly client detection to remove clients from the FL framework, thereby enhancing the security and efficiency of the overall system. This proposed method improves the global model convergence in almost 50\% fewer communication rounds compared with widely used random client selection using the MNIST dataset.

\end{abstract}
\section{Introduction}
The notion of federated learning (FL) was introduced in 2016 \cite{McMahan2017}. Its core idea is to train machine learning models on independent datasets dispersed across multiple devices or parties, preserving local data privacy to some level. Since then, FL has grown rapidly and become a popular research area in the field of artificial intelligence \cite{Ng2020}. The progress is primarily driven by three factors: the widespread use of machine learning technology, the rapid rise of big data, and global data privacy legislation.
\begin{figure}[ht]
\includegraphics[scale=0.6]{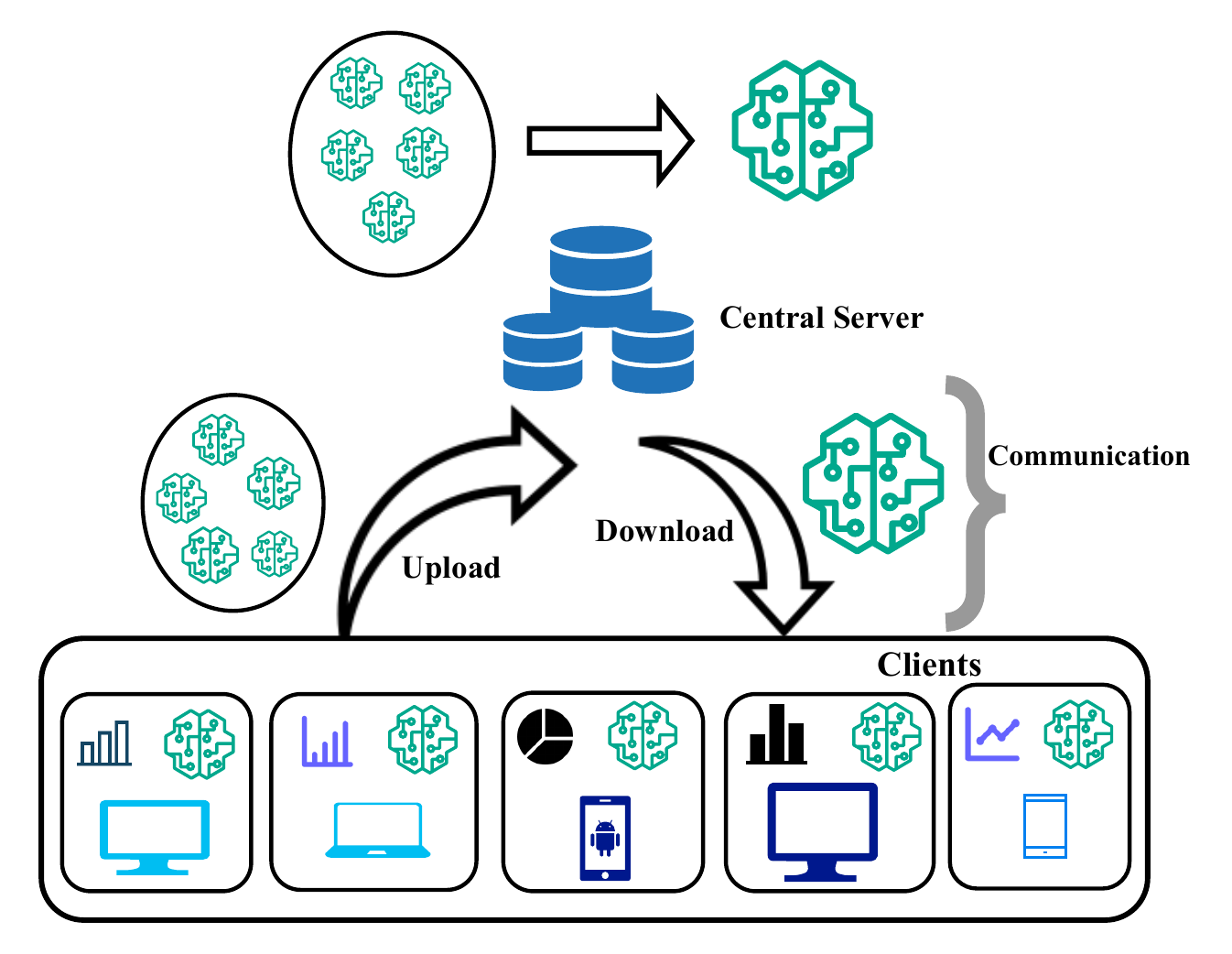}
\caption{Traditional FL Framework}
\label{fig:fl}
\end{figure}
Hence, security and privacy are major characteristics of federated learning. For example, FL is susceptible to Byzantine attacks (which try to stop the model from converging) and poisoning attacks (which try to force convergence to an inaccurate model). FL is particularly vulnerable to Byzantine attacks, in which malicious users (Clients) alter trustworthy models or gradients to obstruct learning or purposely contaminate training data, causing the global model to pick up false information \cite{Zhang2024}. In a traditional FL algorithm as shown in Figure \ref{fig:fl}, initially, the server sends global model parameters to all the randomly selected clients. The clients run their models with their dataset with the received parameters and send the updated parameter values to the server for aggregation. The aggregated parameter values are again sent by the server to the randomly selected clients. Clients vary greatly in terms of hardware configurations and data distribution in a typical FL environment. As a result, each training round's random client sampling may not adequately take advantage of the local updates from heterogeneous clients, which could lead to decreased model accuracy, a slower rate of convergence, compromised fairness, etc. Many client selection methods have been proposed to address the FL client heterogeneity problem, with promising performance improvements. However, none of the client selection methods emphasized identifying the anomalous client and refraining them from participating in the FL network which may also lead to decreased model accuracy, a slower rate of convergence, compromised fairness, and even sabotage the whole FL network. As a solution, we can include another server for maintaining client-server binding or running any privacy-security algorithm to identify the anomalous client, it will enhance the communication and computation complexity of the FL network, which researchers want to minimize for sustainable AI. An anomalous client can be one of the major privacy leakages and (or) security threat scenarios \cite{Yin2021}.

Three scenarios could result in privacy leaking and (or) security threats if a client is anomalous. Initially, the aggregator can provide the anomalous client with intermediate training updates, which they can use to examine confidential data from other client datasets \cite{bhowmick2019protection}. Secondly, the anomalous client may submit training updates to the aggregator that are specifically tailored to probe the unique private data of other client datasets. Third, The attacker can be one of the participants in the federated learning, who adversarially modifies his parameter uploads $W_i^{t}$. Although these concerns can be mitigated by creating customized training processes, such as choosing a subset of customers for each round, FL is still exposed to a significant risk of anomalous clients leaking confidential information or any other security threat. The client can be malicious or anomalous depending on the type of attack \cite{Giuseppe2021}. It is pertinent to mention that whatever the type of attack it is essential to identify the malicious or anomalous clients and refrain them from taking part in the global aggregation because malicious clients can effortlessly incorporate backdoors into the combined model, all the while preserving the model's functionality for the primary goal. Moreover, the presence of anomalous clients may render model poisoning. Model poisoning, as opposed to data poisoning, entails a hostile client going for the global model directly \cite{Zhou2021,Bhagoji2018}. According to research by Bhagoji et al. \cite{Bhagoji2018}, model positioning attacks have a far greater impact on the model than data poisoning attacks. The modifications made to the client's model by the malicious client, which are frequently made in FL before the new model being uploaded to the central server\cite{Cao2019}, have a direct impact on the global model, as illustrated in Figure \ref{fig:attackedfl}. Model poisoning affects the performance of the global model and lowers its overall accuracy by manipulating local model gradients using gradient manipulation technologies \cite{Jere2021,Hong2020}. For example, in Li et al. \cite{Li2021}, when utilizing FL for image recognition, the classifier of an image model may be changed so that it assigns labels chosen by the attacker to specific areas of the image. By teaching rule modification techniques to provide attackers access to the trained model, model poisoning can also be achieved. The attacker can make their attack undetectable by changing the model's output, which allows the trained model to be updated normally \cite{Jere2021,Kairouz2021}. Wahab et al. \cite{Wahab2021} added penalty terms to reduce the discrepancies between the objective functions and appropriate weight update distribution. This enhancement led to the successful deployment of an undetected targeted model poisoning. These motivate us to propose a secure FL algorithm. 
\par
The proposed FL algorithm not only identifies the anomalous client based on the anomaly score, but this algorithm will also refrain the anomalous clients from participating in global averaging. \\
The main contributions are as follows:\\
\begin{itemize}
    \item We propose a secure federated averaging algorithm to detect anomalous clients. Then refrain from those anomalous clients for further participation.
    \item The proposed algorithm is compared with the traditional FedAvg algorithm to validate its performance using iid, non-iid, and non-iid with an unequal number of features.
    \item To validate the proposed concept MNIST dataset of handwritten digits is used for the experiment. The proposed algorithm is secure in terms of client selection and assures convergence earlier by 50\%. Hence, reduces the communication rounds. 
  
\end{itemize}

\section{Related Work}

In the literature, there is substantial evidence to support the idea that federated learning frameworks enhance the security and privacy of data and networks. Furthermore, pertinent research has shown that specific characteristics of a client's private data can be revealed by models or updates provided by the client \cite{Enthoven2021}. With the use of these methods, a malicious server can ascertain whether a particular data point or a sample of data points from the training distribution is used during training or not (reconstruction attacks). For instance, Melis et al. \cite{Melis2018ExploitingUF}
discovered that during training, shared gradients can be used to collect membership data and unintentional feature leakages. Many strategies have been put forth to combat the security and privacy anomalies in FL and guard against such attacks \cite{Xie2018ZenoDS,MA2022103561}. Blanchard et al. \cite{NIPS2017_f4b9ec30}, examined for client changes during each loop in the context of security assaults and eliminated potentially dangerous clients when the server gathered updates. To protect against reconfiguration attacks, Gao et al. \cite{Gao2021} suggested looking for privacy-preserving transformation functions and preprocessing training samples with them in order to guarantee the trained model performs well. In \cite{Liu2022FederatedLW}, the authors investigated the resilience of federated learning and suggested the PUS-FL federated learning parameter update architecture based on blockchain technology. The authors demonstrated through experiments that simulate distributed machine learning on neural networks that PUS-FL's anomaly detection algorithm performs better than traditional gradient filters, such as geometric median, Multi-Krum and trimmed mean. Furthermore, it confirms that the scalability-enhanced parameter aggregation consensus algorithm (SE-PBFT) presented in \cite{Liu2022FederatedLW} reduces communication complexity to increase consensus scalability. 
However, very few of them discussed anomalous client detection in a way that prevents anomalous clients from participating in aggregation and causing the model to fail to converge.

\section{Proposed Algorithm}
The proposed method is based on the traditional FedAvg algorithm \cite{McMahan2017}. Here we compare traditional FedAvg \cite{McMahan2017} with our proposed secure FL algorithm.
The Algorithm \ref{algo:fedavg} presented the traditional FedAvg algorithm. 

\begin{algorithm}[ht]
\caption{Traditional Federated Averaging Algorithm}\label{algo:fedavg}
\begin{algorithmic}[1]
\Require
\State  {$C$ $\gets$ number of clients indexed by $c$}
\State {$E$ $\gets$ number of local epochs}
\State {$B$ $\gets$ local minibatch size}
\State {$\eta$ $\gets$ learning rate}
\State {$D_c$ $\gets$ dataset at client $c$}
\NoNumber{\textbf{Server Execution:}}
\State {Initialize $w_0$}
\For {each round $t=1,2,\cdots$}
\State $m \gets max(\lfloor{F.C \rfloor}, 1)$ \Comment{F controls the global batch size}
\State {$s_t$= random set of $m$ clients}
\For {each client $c \in S_t$ in parallel}
 \State {$w_{t+1}^c$ $\gets$  ClientUpdate$(c,w_t)$}
 \EndFor
 \State {$m_t \gets \sum_{c \in S_t}n_c$}
 \State {$w_{t+1} \gets \sum_{c \in S_t}\frac{n_c}{m_t} w_{t+1}^c$}
 \EndFor
\State{\NoNumber{\textbf{Client Execution:}}}
\State {ClientUpdate $(c,w_t)$:} \Comment{Each client, $c$ executes}
\State {$\mathcal{B}$ $\gets$ (split $D_c$ into batches of size $B$)}
\For {each local epoch i from 1 to E}
\For {batch $b \in \mathcal{B}$}
\State {$w \gets w - \eta \nabla{l(w ; b)}$}
\EndFor
\EndFor 
\State {return $w$ to the server}
\end{algorithmic}
\end{algorithm}
\begin{figure}[ht]
\includegraphics[scale=0.6]{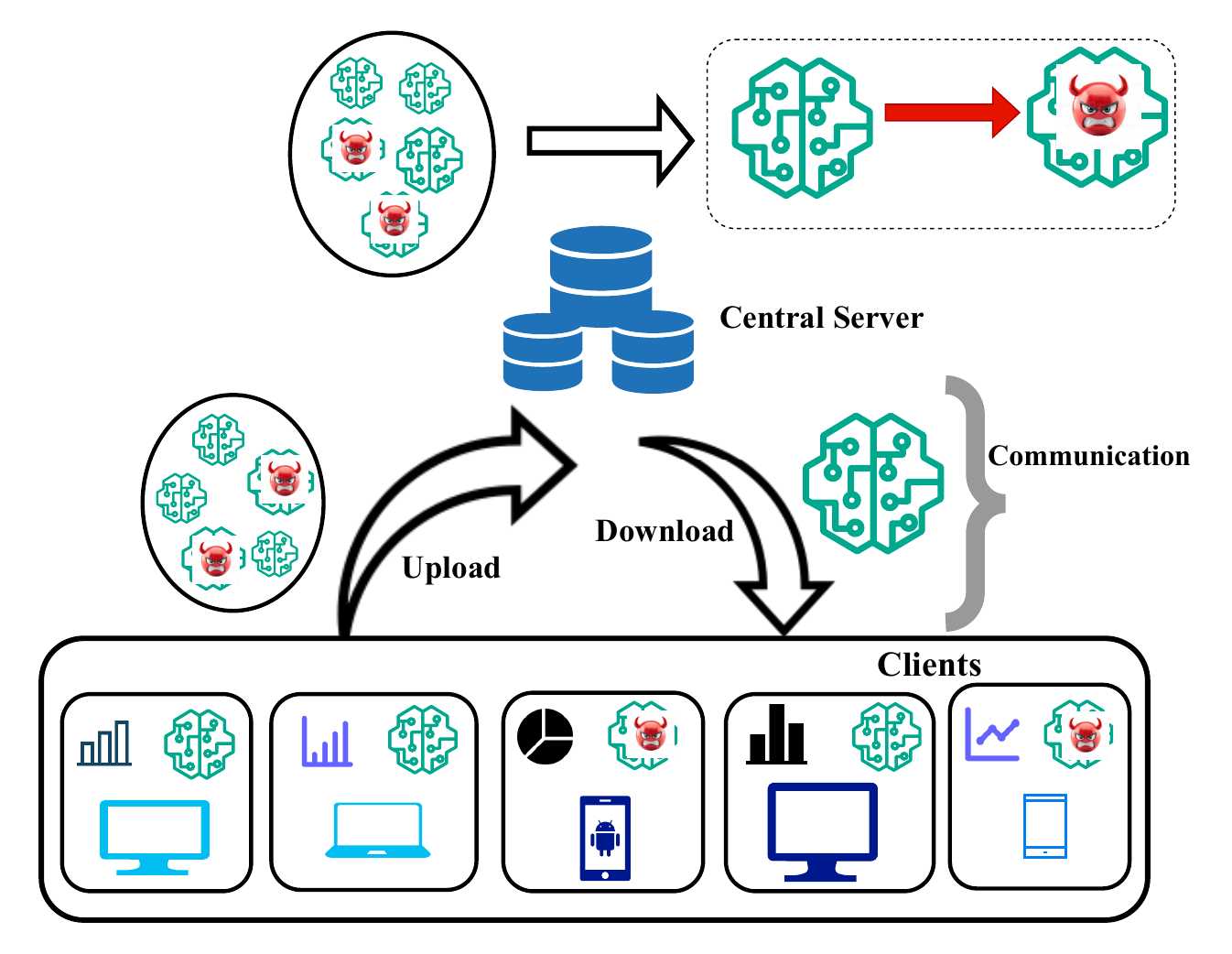}
\caption{Attacked FL Framework with Anomalous Client}
\label{fig:attackedfl}
\end{figure}
The traditional FedAvg algorithm uses a central server to coordinate training. This server houses the shared global model $w_t$, where $t$ represents the communication round. Nevertheless, the real optimization is carried out locally on clients with the help of Stochastic Gradient Decent (SGD). The fraction of clients $F$ to choose for training, the local mini-batch size $B$, the number of local epochs $E$, a learning rate $\eta$, and potentially a learning rate decay $\lambda$ are the five hyperparameters of FedAvg. When using SGD for training, the parameters $B$, $E$, $\eta$, and $\lambda$ are frequently utilized. But in this case, $E$ represents the total number of iterations through the local data prior to an update to the global model. Initially, the server sends the global model, $w_0$ to all the randomly selected clients. Each communication round in FedAvg includes the following steps: The server delivers the current global model $w_t$ to every client in $S_t$ after choosing a subset of clients $S_t$, $|S_t| = F · C \ge 1$. Each client divides their local data into batches of size $B$ and executes $E$ epochs of SGD after updating their local models, $w_t^c$, to the shared model, $w_t^c \gets w_t$. Lastly, clients upload their trained local models $(w_t^k+1)$ to the server, which creates the new global model $(w_t+1)$ by adding up all of the local models that were received. The number of local training instances determines the weighting system, as stated in Algorithm \ref{algo:fedavg} on lines 12 and 13.

\begin{algorithm}[ht]
\caption{Proposed Secure Federated Averaging Algorithm}\label{algo:securefedavg}
\begin{algorithmic}[1]
\Require 
\State  {$C$ $\gets$ number of clients indexed by $c$}
\State {$E$ $\gets$ number of local epochs}
\State {$B$ $\gets$ local minibatch size}
\State {$\eta$ $\gets$ learning rate}
\State {$D_c$ $\gets$ dataset at client $c$}
\State {$A^{th}$ $\gets$ threshold of anomaly score}
\NoNumber{\textbf{Server Execution:}}
\State {Initialize $w_0$, $A_0=1$, $S_0={\phi}$} \Comment{$w_0$ is the global weight and $A_0$ is the anomaly score}
\If{round $t=1$}
\State $m \gets max(\lfloor{F.C \rfloor}, 1)$ \Comment{F controls the global batch size}
\State {$s_1$= random set of $m$ clients}
\EndIf
\For {each round $t=2,3\cdots$}
 \State{$\{S_t\} = \{s_1\} - \{S_a\} $}
\For {each client $c \in S_t$ in parallel}
 \State {$w_{t+1}^c$ $\gets$  ClientUpdate$(A_t,c,w_t)$}
 \State {$A_{t+1}^c$ $\gets$ $\frac{1+loss(w_{t+1}^c)}{1+\sigma_{t+1}}$}
 \State{$\sigma_{t+1}=min_c \{loss(w_{t+1}^c), c=1,2,\cdots, C \}$}
 \If{$A_{t+1}^c > A^{th}$}
 \State{Add that client index in set $S_a$}
 \EndIf
 \EndFor
 \State {$m_t \gets \sum_{c \in S_t}n_c$}
 \State {$w_{t+1} \gets \sum_{c \in S_t}\frac{n_c}{m_t} w_{t+1}^c$}
 \EndFor

\State{\NoNumber{\textbf{Client Execution:}}}
\State {ClientUpdate $(c,w_t)$:} \Comment{Each client, $c$ executes}
\State {$\mathcal{B}$ $\gets$ (split $D_c$ into batches of size $B$)}
\For {each local epoch i from 1 to E}
\For {batch $b \in \mathcal{B}$}
\State {$w \gets w - \eta \nabla{l(w ; b)}$}
\EndFor
\EndFor 
\State {return $w$ to the server}

\end{algorithmic}
\end{algorithm}

In the proposed secure federated averaging algorithm, we calculate the anomaly score of all the participating clients. If the anomaly score of a client is greater than the anomaly threshold value then that client is restricted to participate. Figure \ref{fig:attackedfl} demonstrates the FL model diagram when clients become anomalous. Even, the client is unable to send the parameter values to the global server for aggregation.  Algorithm \ref{algo:securefedavg} represents the federated version of the proposed method step by step. We initialize the anomaly score as 1, assuming the loss as zero as clients never start their executions. The global model sends $w_0$ to all the randomly selected models in the first round. Then the clients update itself with the global parameters and local iterations. Each client calculates its local loss $\nabla{l(w; b)}$ based on their local datasets. In the further rounds, the anomaly score is calculated based on the local loss and the minimum loss of all the participating clients using the formula $A_{t+1}^c$ $\gets$ $\frac{1+loss(w_{t+1}^c)}{1+\sigma_{t+1}}$ where $\sigma_{t+1}=min_c \{loss(w_{t+1}^c), c=1,2,\cdots, C \}$. Finally, after the completion of the local updates each client compares its calculated anomaly score with the threshold anomaly score. If any of the client's anomaly score is greater than the threshold then that client will be treated as an anomalous client and that client will be removed from the randomized selected client set. 
 The threshold $A^{th}_{t+1}$ can be chosen as, e.g., the average value or the median of the anomaly scores $\{A_{t+1}^c, c=1,2,\cdots, C \}$.
In such a way, we can detect the anomalous client and remove it from the set of participating clients.

\section{Experiment}
\subsection{Experimental Setup}


We evaluate our proposed method with the MNIST dataset of handwritten digits, which has a training set of 60,000 examples, and a test set of 10,000 examples. It is a subset of a larger set available from NIST. The digits have been size-normalized and centered in a fixed-size image. In the FL settings, we divide the dataset in 3 different ways such as iid, and non-iid with equal number of features and non-iid with unequal number of features. It is pertinent to mention that all the clients have received the dataset as first we assign each client one shard to ensure every client has one shard of data. Next, the remaining shards are assigned randomly. 
\begin{figure}[!h]
     \centering
     \begin{subfigure}[b]{0.4\textwidth}
         \centering
         \includegraphics[width=\textwidth]{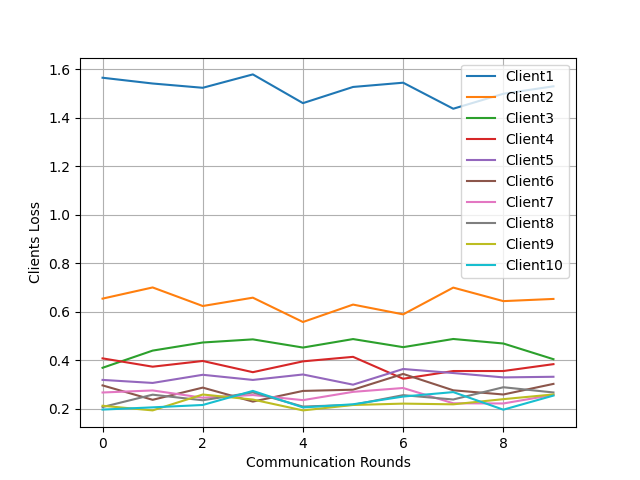}
         \caption{Loss of each clients with iid dataset with anomalous clients}
    \label{fig:clientloss}
     \end{subfigure}
     \hfill
     \begin{subfigure}[b]{0.4\textwidth}
         \centering
         \includegraphics[width=\textwidth]{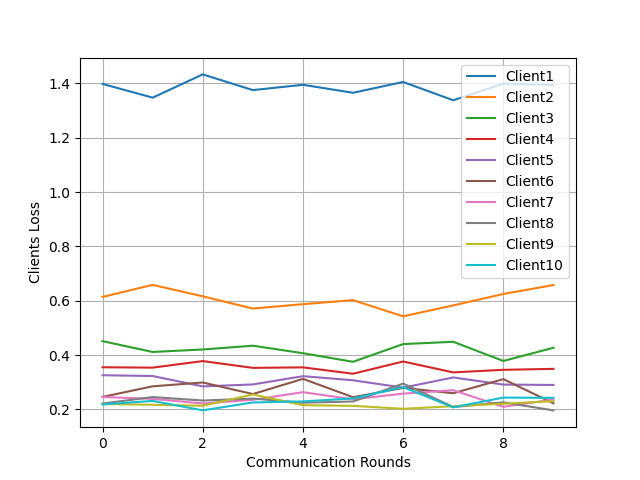}
          \caption{Loss of each clients with non-iid dataset with equal number of features with anomalous clients}
    \label{fig:clientlossiid}
     \end{subfigure}
     \hfill
     \begin{subfigure}[b]{0.4\textwidth}
         \centering
         \includegraphics[width=\textwidth]{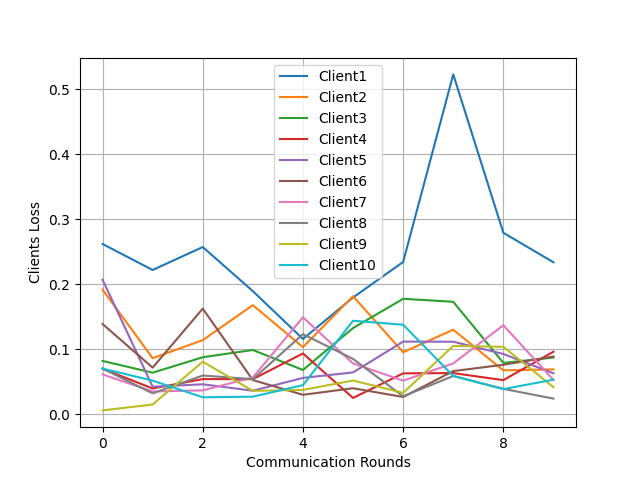}
         \caption{Loss of each clients with non-iid dataset with unequal number of features with anomalous clients}
    \label{fig:clientlossnoniid}
     \end{subfigure}
        \caption{Clients loss w.r.t communication rounds with anomalous clients}
        \label{fig:clients loss}
\end{figure}

A CNN model is created using PyTorch to train with the datasets. The CNN model with two convolutional layers, each with a 5x5 kernel, 32 channels, and 2x2 max-pooling, is the image classification model taught in federated learning. It is succeeded by a fully connected layer with 1024 units and a 62-dimensional softmax output layer. Within the concealed layers, ReLu activation is employed. The dataset is divided among 10 different clients. The FL was carried out for 10 communication rounds with 10 epochs in each communication round. The batch size is considered as 124 for all the experiments and the learning rate is 0.1. The parameters are fixed for all the experiments performed using the traditional federated averaging algorithm (FedAvg) and our proposed secure FL algorithm.
\begin{figure}[!h]
     \centering
     \begin{subfigure}[b]{0.4\textwidth}
         \centering
         \includegraphics[width=\textwidth]{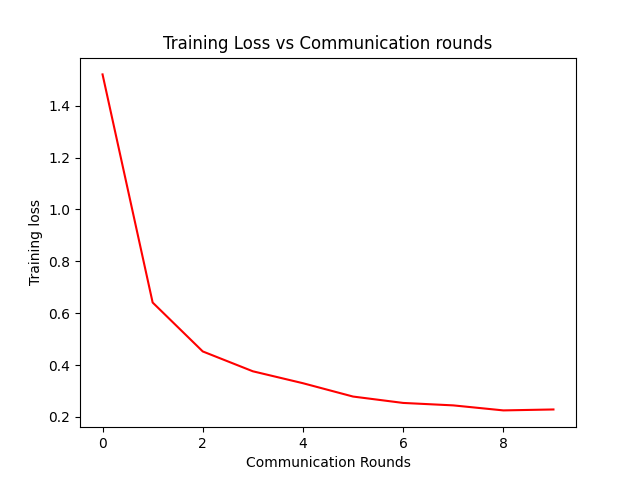}
         \caption{Global loss with iid dataset without anomalous clients}
    \label{fig:gloveloss}
     \end{subfigure}
     \hfill
     \begin{subfigure}[b]{0.4\textwidth}
         \centering
         \includegraphics[width=\textwidth]{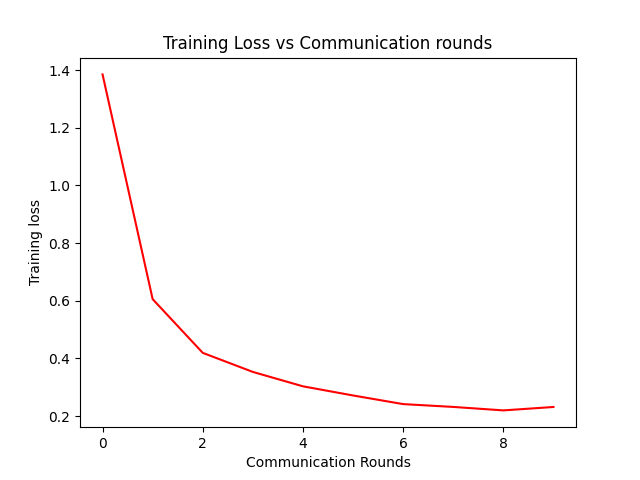}
          \caption{Global loss with non-iid dataset with equal number of features without anomalous clients}
    \label{fig:glovelossiid}
     \end{subfigure}
     \hfill
     \begin{subfigure}[b]{0.4\textwidth}
         \centering
         \includegraphics[width=\textwidth]{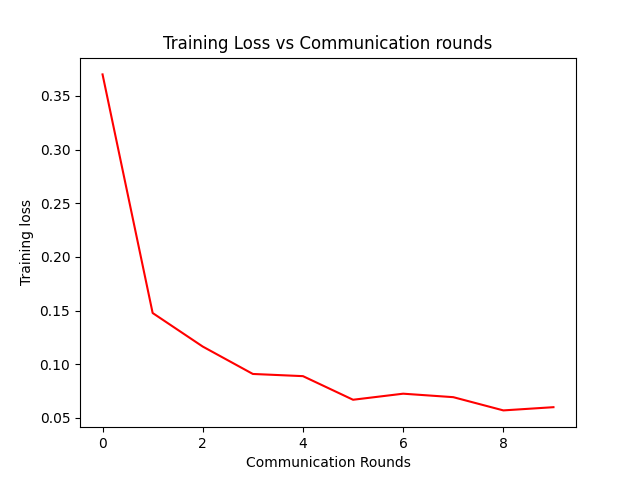}
         \caption{Global loss with non-iid dataset with unequal number of features without anomalous clients}
    \label{fig:globelossnoniid}
     \end{subfigure}
        \caption{ Global Loss w.r.t communication rounds without anomalous clients}
        \label{fig:Global loss}
\end{figure}

\subsection{Experimental Results}

The real weights of the first layer of the ML model are supplemented with a small random seed in order to generate anomalous clients. The Figures \ref{fig:clients loss} represent converging scenarios after manually changing the weight of the 1st layer of the two clients. From the Figures \ref{fig:clients loss}, we can clearly understand that client 1 and client 2 behave abnormally. If the global model never considers those clients in the aggregation then the global model started converging after 4 communication rounds as shown in the Figure \ref{fig:Global loss} for 3 different distributions of the dataset. However, the real convergence of the global model with a non-iid dataset with an unequal number of features started after 8 communication rounds. We also compare the global accuracy with the FedAvg \cite{McMahan2017} algorithm with three different distributed datasets as shown in Table \ref{tab:tab1} after 10 global rounds.
\begin{table}
\centering
    \begin{tabular}{|c|c|c|}
    \hline
        FL Algorithm & Data Distribution & Accuracy \\
        \hline
        \multirow{3}{*}{FedAvg} & IID & 95.12\% \\
         & Non-IID &   92.34\%\\
         & Non-IID with unequal number of features & 86.04\%\\
         \hline
         \multirow{3}{*}{Proposed Secure FL} & IID & 98.54\% \\
         & Non-IID &  98.02\% \\
         & Non-IID with unequal number of features &  96.45\%\\
         \hline
    \end{tabular}
    
    \caption{Comparison of our proposed algorithm with the FedAvg algorithm with CNN after 10 global rounds}
    \label{tab:tab1}
\end{table}
From Table \ref{tab:tab1}, it is clear that our proposed algorithm is converging early with higher accuracy in comparison with traditional FedAvg. To get the 99\% accuracy with traditional FedAvg, it is required to run more than 100 global iterations according to the literature. However, our algorithm converges early as the selected clients are performing we in their local iterations as we remove the anomalous clients.

\section{Conclusion}
 In FL, data remains on edge devices for local processing. Federated learning (FL) has several advantages over centralized machine learning (ML), including edge computing, enhanced privacy, and reduced server costs. Since the ML training process involves multiple edge devices, it's crucial to understand the state of each device. To identify anomalous clients in a distributed system, a federated secure algorithm is introduced in this paper. The main concept of this algorithm is to calculate the anomaly score of each local model based on the local loss and the minimum loss produced by all local models. Then, the algorithm uses a threshold anomaly score to identify abnormal clients, removing them from the client set for subsequent iterations. This not only detects anomalous clients but also provides a secure client selection algorithm in FL settings. The proposed method is tested on the MNIST image classification dataset, and in the future, it will be further tested with more clients using other domain-specific datasets.
\begin{credits}
\subsubsection{\ackname} This work contributes to the basic research activities of the WP9.6: "AI for Green" supported by the PNRR project FAIR -  Future AI Research (PE00000013), Spoke 9 - Green-aware AI, under the NRRP MUR program funded by the NextGenerationEU.

\subsubsection{\discintname}
The authors have no competing interests to declare that are
relevant to the content of this article.
\end{credits}
%
%
%
 \bibliographystyle{splncs04}
 \bibliography{mybibfile}

\end{document}